# Supervised Multi-Modal Fission Learning

Lingchao Mao[a], Qi Wang[b], Yi Su[c], Fleming Lure[d], Jing Li[a]

*[a]H. Milton Stewart School of Industrial and Systems Engineering, Georgia Institute of Technology, Atlanta, GA, USA; [b]ASU-Banner Neurodegenerative Disease Research Center, Arizona State University, Tempe, AZ, USA; [c]Banner Alzheimer's Institute, Phoenix, AZ, USA; MS Technologies Corporation, Rockville, MD, USA*

*Corresponding author: Jing Li (jli3175@gatech.edu)

**Abstract**

Learning from multi-modal datasets can leverage complementary information and lead to improved performance for prediction tasks. To account for feature correlations in high-dimensional datasets, a commonly used strategy is the latent variable approach. Several latent variable methods in the literature have been proposed for multi-modal datasets, however, these methods either focus on extracting the shared component across all modalities or extracting a shared component and individual components specific to each modality. To address this gap, we propose a Multi-Modal Fission Learning (MMFL) model that simultaneously identifies globally joint, partially joint, and individual components underlying the features of multi-modal datasets. Unlike existing latent variable methods, MMFL uses supervision from the response variable to identify predictive latent components and has a natural extension to incorporate incomplete multi-modal data. Through simulation studies, we demonstrate that MMFL outperforms a variety of existing multi-modal algorithms under both complete modality and incomplete modality settings. We applied MMFL to a real-world case study for early prediction of Alzheimer's Disease using multi-modal neuroimaging (MRI and PET) and genetic data (SNP) from the Alzheimer's Disease Neuroimaging Initiative (ADNI) dataset. MMFL provided more accurate predictions and better insights for understanding within and across modality correlations compared to existing methods.

## 1. Introduction

Multi-modal datasets widely exist in various science and engineering domains. The different modalities provide complementary information, leading to improved performance for prediction tasks. For instance, in health and medicine, it is typical to have multi-modal datasets collected from each patient for diagnosis and prognosis purposes, including various types of medical imaging (MRI, PET), genetics, and others, especially for complicated diseases such as Alzheimer's Disease (AD) and cancer (Cartwright, 2021; Pang et al., 2021). Developing machine learning (ML) models to effectively and efficiently integrate the complex, multi-modal datasets is critical for providing accurate diagnosis or prognosis for each patient.

To integrate multi-modal datasets to build a predictive model, an intuitive approach is to concatenate features from the different modalities into a single data matrix, which can be subsequently used to train any supervised learning model (Abdi et al., 2013; Chen et al., 2009; Smilde et al., 2003). This approach is easy to implement and allows practitioners to quickly adopt off-the-shelf ML packages for their datasets. The downside is that the concatenated feature vector may contain substantial redundancy due to the correlations of features within each modality and between modalities. Another common strategy is to train modality-specific models and integrate their respective predictions or extracted features. Ensemble learning (Sagi & Rokach, 2018), Multiple Kernel Learning (MKL) (Hinrichs et al., 2009), and co-training (Blum & Mitchell, 1998; Zhu et al., 2014) methods are along this line. While this approach enjoys the benefit of retaining inherent structure of multi-modal data and accounting for within-modality feature correlations, it cannot effectively account for feature correlations between different modalities as features are capsulated within each modality. This leads to potential redundancy and lose of flexibility in modelling feature-level relationship with the response variable.

To account for feature correlations, latent variable approaches are commonly used, which assume the existence of latent variables underlying the observed features that explain the correlations of the observed features. Classical examples include Principal Component Analysis (PCA) (Wold et al., 1987), Factor Analysis (FA) (Kim & Mueller, 1978), Non-negative Matrix Factorization (NMF) (Lee & Seung, 2000), and Canonical Correlation Analysis (CCA) (Thompson, 1984). These approaches were originally invented for single-modality datasets. Later, with the prevalence of multi-modal datasets in various application domains, these approaches have been extended to multi-modal settings, resulting in methods such as Joint and Individual Variation Explained (JIVE) (Lock et al., 2013) and Structural Learning and Integrative Decomposition (SLIDE) (Gaynanova & Li, 2019). However, these methods are unsupervised, meaning that they do not have a response variable to supervise the process of identifying the latent variables (a.k.a. components in many of these methods). As a result, even though the identified components facilitate a better understanding of the correlation structure of multi-modal datasets and enable dimension reduction, they are not necessarily good predictors for the response variable.

More recently, researchers have started to interrogate the question as to how to modify latent variable approaches of multi-modal datasets to suit the supervised learning setting. This led to the development of methods such as supervised JIVE (sJIVE) (Palzer et al., 2022) and Complete Multi-modal Latent Space (CMLS) learning model (T. Zhou et al., 2020). However, existing methods either focus on extracting the shared component across all modalities to be linked with the response variable or extracting shared components and individual components specific to each modality. The former lacks a consideration that there may be information specific to each modality that also helps with the prediction. The latter embraces this consideration and provides an adequate approach for two-modality cases. However, if there are more than two modalities, there could be components shared by subsets of the modalities (namely, partially shared components), in addition to the globally shared and individual components. This more general situation cannot be effectively accounted for by existing methods. This motivated us to develop a new method that can identify globally shared, partially shared, and individual components from multi-modal datasets beyond two modalities in the supervised learning setting, namely the multi-modal fission learning (MMFL) model. Here, we adopt the word "fission" from nuclear physics, which describes the splitting of atoms to release energy, to mean the splitting of multi-modal data into succinct, non-overlapping latent components to best serve the prediction task.

Additionally, it is common for a sample not to possess all the modalities. For instance, in health and medicine, some modalities may not be covered by the insurance policy of the patient and is too expensive if paying out of pocket; some modalities may not be available in the clinic or hospital where the patient receives care; some modalities may be prohibited for some patients due to safety concerns. Similarly, in engineering applications, missing sensor signals can be common as continuous observations is not economical (Fang et al., 2015). Thus, it is important to have the capability to integrate incomplete multi-modal datasets in training as well as at inference time. To our best knowledge, limited multi-modal latent variable approaches have this capability.

The main contributions of this article are as follows:

(1) New model formulation: we propose an MMFL model that simultaneously identifies globally joint, partially joint, and individual components underlying the features of multi-modal datasets in a supervised manner. MMFL extracts non-redundant, complementary latent components from multi-modal datasets for prediction, satisfying the consensus and complementary principles of multi-modal learning.

(2) Extension to address incomplete multi-modal datasets: We propose an extension of MMFL that can be used for the incomplete modality settings. This was accomplished by leveraging the modality structure assumptions to estimate pseudo-reconstructions for the missing modalities. Our experiments demonstrated that incomplete modality learning was more advantageous compared to learning using only samples with non-missing modalities or using only modalities with non-missing samples.

(3) Comparison with existing multi-modal learning algorithms: We evaluated the performance of our proposed MMFL model under both complete and incomplete modality settings. MMFL showed significantly better predictive performance compared to various existing multi-modal algorithms.

(4) Application in early prediction of AD: We applied MMFL to a real-world case study_for predicting the conversion to AD for individuals with Mild Cognitive Impairment (MCI) using multi-modal neuroimaging and genetic data from the Alzheimer's Disease Neuroimaging Initiative (ADNI). MMFL provided reasonably accurate prediction accuracy along with insights for understanding correlations within and across imaging and genetic modalities.

The remainder of this paper is organized as follows. Section 2 provides an overview of related works. Section 3 presents the methodological details of the proposed MMFL model. Section 4 validates the proposed method by comparing against existing methods using simulation data. Section 5 presents a real-world case study for early prediction of AD. Lastly, we conclude the paper with a brief discussion and outlook for future work in Section 6.

## 2. Related work

Various ML models have been developed to integrate multi-modal datasets in prediction tasks. At a high level, the existing methods can be divided into those based on deep neural networks (DNN) and those based on other non-DNN ML models. While recognizing the importance of the former field, the proposed method belongs to the latter field which are easier to interpret and less prone to overfitting, especially for problems with a smaller training set. Since the methods in the two fields are quite different in their designs, we focus on reviewing the relevant works in the latter field which our proposed method belongs to, while referring interested readers to recent review papers on the former field such as (Ramachandram & Taylor, 2017).

Latent variable approaches have been an effective strategy to handle large-dimensional datasets with capability to account for feature correlations. The essence of latent variable methods lies in assuming the existence of lower-dimensional latent subspace from which the observed variables are generated, which are called latent variables, components, or factors. Generally, the given data matrix $\boldsymbol{X}$ is factorized into $\boldsymbol{X} = \boldsymbol{U}\boldsymbol{V}^T$, where $\boldsymbol{U}$ is the matrix of latent variables, and $\boldsymbol{V}$ is the loading matrix. However, finding the latent subspace is an ill-posed problem as any invertible matrix $\boldsymbol{T}$ can be applied to transform $\boldsymbol{U}$ and $\boldsymbol{V}$ such that $\boldsymbol{X} = (\boldsymbol{U}\boldsymbol{T}^{-1})(\boldsymbol{T}\boldsymbol{V}^T)$. A unique factorization can be established by imposing sufficiently strong constraints, such as orthogonality (Wold et al., 1987), nonnegativity (Lee & Seung, 2000), physically meaningful factors (Kim & Mueller, 1978), independence (Lee, 1998), correlation (Thompson, 1984), low-rank structure (Kolda & Bader, 2009), smoothness (C. Zhang et al., 2017), or diversity (Cao et al., 2015). These methods have been extended to multi-modal settings, resulting in methods such as integrated PCA (Tang & Allen, 2021), group factor analysis (Klami et al., 2015), multi-view CCA (Rupnik & Taylor, 2010), sparse multi-view matrix factorization (Z. Wang et al., 2015), group NMF (Peng et al., 2022; Yang & Michailidis, 2016), and their variants.

An important constraint to consider in the multi-modal setting is the modality structure. The simplest structure is to assume that different modalities are generated from a common latent space. A straightforward example is applying PCA to the concatenated data matrix (Smilde et al., 2003). Other works such as multi-view CCA (Rupnik & Taylor, 2010), multi-view Fisher Discriminant Analysis (Diethe et al., 2008), and joint or group NMF (Peng et al., 2022; S. Zhang et al., 2012) also focus on extracting the common components across different datasets. However, assuming all modalities are generated from the same latent subspace is a too strong assumption. To relax this assumption, Joint and Individual Variation Explained (JIVE) (Lock et al., 2013) constructed each modality from two latent spaces: a joint latent space shared across all modalities and an individual latent space specific to the modality. Subsequent work are also along this direction (Argelaguet et al., 2018; Feng et al., 2018; Peng et al., 2022; Ray et al., 2014; Risk & Gaynanova, 2021; Schouteden et al., 2013; Tang & Allen, 2021; Yang & Michailidis, 2016; Zhou et al., 2016). This approach adequately considers within and across modality correlations for two-modality cases. However, if there are more than two modalities, they are not able to reveal insights about correlations among subsets of modalities which are originated from partially shared components.

A more general formulation for multi-modality data was proposed in the Structural Learning and Integrative Decomposition (SLIDE) model (Gaynanova & Li, 2019), where data is factorized into globally-shared, partially-shared, and individual components. The authors demonstrated that incorporating the partially-joint components not only improved predictive performance but also allowed better understanding of cross-modality correlations (Gaynanova & Li, 2019). Following this work, (Choi & Jung, 2022) attempted to improve the estimation of partial components and (H. Chen et al., 2022) proposed an adapted version for four-dataset setting. However, these methods are still unsupervised, meaning that the latent components they identify accounts for the correlation structure of multi-modal datasets but are not necessarily good predictors for the response.

More recently, researchers started to develop models that employ supervision to strengthen the discriminative ability of the learnt latent components to suit the multi-modal supervised learning setting. Earlier works posed a two-step design (Kaplan & Lock, 2017), introduced covariates (Gao et al., 2021; G. Li & Jung, 2017), used the label as an additional modality (Rodosthenous et al., 2020), or maximized correlation with the response variable (Safo et al., 2022; Y. Zhang & Gaynanova, 2022). The CMLS model (T. Zhou et al., 2020) was one of the first to include an explicit supervised loss while learning a projection of all modalities into a common latent space. However, these works still focused on extracting signals that is shared by all modalities, which ignores information specific to each modality that may help with the prediction. To embrace this consideration, sJIVE extended JIVE to the supervised setting (Palzer et al., 2022). Similarly, (P. Wang et al., 2022; Xue et al., 2017) proposed models that extracted globally joint and individual components for prediction. However, these models cannot extract partially shared components between subsets of modalities if there are more than two modalities. This motivated us to develop MMFL, a new supervised multi-modal learning model that can identify globally shared, partially shared, and individual components from multi-modal datasets beyond two modalities.

Additionally, there is a lack of consideration for incomplete modalities in the existing multi-modal latent variable learning approaches. In various applications where data collection is expensive or not always possible, it is common that a sample only has partial data collected. Samples with partial modalities or modalities with very few samples are either discarded or imputed in the preprocessing step before model training. Given the scarcity and complexity of data, it is desirable that the model can still train and predict using incomplete-modality samples. Thus, we extend the proposed MMFL model to allow incomplete modality data in both training and test phases.

# 3. Development of Multi-modal Fission Learning (MMFL) model

In this section, we introduce MMFL as a supervised latent variable approach that learns to decompose multi-modal dataset into globally joint, partially joint, and individual components. We provide optimization algorithms to train MMFL in both complete- and incomplete modality settings. To determine component ranks, we propose two model-agnostic strategies. Lastly, we discuss the design of MMFL with respect to the complementary and consensus principles of multi-modal learning.

## 3.1 Mathematical formulation of MMFL

Consider a multi-modal dataset of $n$ samples and $m$ modalities $\boldsymbol{X} = [\boldsymbol{X}_1, \dots, \boldsymbol{X}_m]$, where $\boldsymbol{X}_i \in \mathbb{R}^{n\times p_i}$ denotes the data matrix for the $i$-th modality with $p_i$ features and $p = \sum_{i=1}^{m} p_i$ is the total number of features. Let $\boldsymbol{y}$ denote the vector of labels associated with the $n$ samples. In binary classification, $\boldsymbol{y} \in \{0,1\}^{n\times 1}$, and in regression, $\boldsymbol{y} \in \mathbb{R}^n$. Assume that the latent components underlying the multi-modal dataset include (1) a globally joint component, $\boldsymbol{U}_g$, that is shared by all the modalities, i.e., $g = \{1, \dots, m\}$; (2) partially joint components that are shared by subsets of the modalities, i.e., $\{\boldsymbol{U}_{\varphi\subset g}\}$; and (3) individual components that are unique to each modality, i.e., $\{\boldsymbol{U}_{i\in g}\}$. Let $\boldsymbol{U} \in \mathbb{R}^{n\times r}$ denote the collection of all the components in (1)-(3), i.e., $\boldsymbol{U} = \left[\boldsymbol{U}_g, \{\boldsymbol{U}_{\varphi\subset g}\}, \{\boldsymbol{U}_{i\in g}\}\right]$. $\boldsymbol{U}$ is a low-rank representation of the multi-modal features, so $r \ll p$.

Then, the relationship between $\boldsymbol{U}$ and the $i$-th modality can be expressed as:

$$\boldsymbol{X}_i = \boldsymbol{U}\boldsymbol{V}_i^T + \boldsymbol{E}_i, \tag{1}$$

where $\boldsymbol{V}_i \in \mathbb{R}^{p_i \times r}$ denotes the loading coefficient matrix and $\boldsymbol{E}_i$ is the noise matrix of the $i$-th modality. It is important to note that not all the components in $\boldsymbol{U}$ contribute to $\boldsymbol{X}_i$, but only the globally shared component, the partially joint components that involve the $i$-th modality, and the individual component that is unique to $i$-th modality. This structure can be enforced by setting the loading coefficients corresponding to non-contributing components to be zero.

To explain this better, use a three-modality case as an example. That is, $\boldsymbol{X} = [\boldsymbol{X}_1, \boldsymbol{X}_2, \boldsymbol{X}_3]$ and $\boldsymbol{U} = \left[\boldsymbol{U}_g, \boldsymbol{U}_{(1,2)}, \boldsymbol{U}_{(1,3)}, \boldsymbol{U}_{(2,3)}, \boldsymbol{U}_1, \boldsymbol{U}_2, \boldsymbol{U}_3\right]$. Then,

$$\boldsymbol{X}_1 = \boldsymbol{U}\boldsymbol{V}_1^T + \boldsymbol{E}_1 = \left[\boldsymbol{U}_g, \boldsymbol{U}_{(1,2)}, \boldsymbol{U}_{(1,3)}, \boldsymbol{U}_{(2,3)}, \boldsymbol{U}_1, \boldsymbol{U}_2, \boldsymbol{U}_3\right] \begin{bmatrix} \boldsymbol{V}_{g,1} \\ \boldsymbol{V}_{(1,2),1} \\ \boldsymbol{V}_{(1,3),1} \\ \boldsymbol{0} \\ \boldsymbol{V}_{1,1} \\ \boldsymbol{0} \\ \boldsymbol{0} \end{bmatrix} + \boldsymbol{E}_1,$$

$$\boldsymbol{X}_2 = \boldsymbol{U}\boldsymbol{V}_2^T + \boldsymbol{E}_2 = \left[\boldsymbol{U}_g, \boldsymbol{U}_{(1,2)}, \boldsymbol{U}_{(1,3)}, \boldsymbol{U}_{(2,3)}, \boldsymbol{U}_1, \boldsymbol{U}_2, \boldsymbol{U}_3\right] \begin{bmatrix} \boldsymbol{V}_{g,2} \\ \boldsymbol{V}_{(1,2),2} \\ \boldsymbol{0} \\ \boldsymbol{V}_{(2,3),2} \\ \boldsymbol{0} \\ \boldsymbol{V}_{2,2} \\ \boldsymbol{0} \end{bmatrix} + \boldsymbol{E}_2,$$

$$\boldsymbol{X}_3 = \boldsymbol{U}\boldsymbol{V}_3^T + \boldsymbol{E}_3 = \left[\boldsymbol{U}_g, \boldsymbol{U}_{(1,2)}, \boldsymbol{U}_{(1,3)}, \boldsymbol{U}_{(2,3)}, \boldsymbol{U}_1, \boldsymbol{U}_2, \boldsymbol{U}_3\right] \begin{bmatrix} \boldsymbol{V}_{g,3} \\ \boldsymbol{0} \\ \boldsymbol{V}_{(1,3),3} \\ \boldsymbol{V}_{(2,3),3} \\ \boldsymbol{0} \\ \boldsymbol{0} \\ \boldsymbol{V}_{3,3} \end{bmatrix} + \boldsymbol{E}_3. \tag{2}$$

Note that the loading coefficients corresponding to non-contributing components to the first modality, i.e., $\boldsymbol{U}_{(2,3)}$, $\boldsymbol{U}_2, \boldsymbol{U}_3$, are set to be zero. The other two modalities can be expressed in a similar way.

To aggregate the modality-wise equation in (1) into a single equation, we can write

$$\boldsymbol{X} = \boldsymbol{U}(\boldsymbol{V} \circ \boldsymbol{S})^T + \boldsymbol{E}, \tag{3}$$

where $\boldsymbol{V} = [\boldsymbol{V}_1^T \dots \boldsymbol{V}_m^T]^T \in \mathbb{R}^{p\times r}$, $\boldsymbol{S}$ is a 0/1 matrix of the same dimension as $\boldsymbol{V}$ using zeros to indicate non-contributing components to each modality, $\circ$ denotes element-wise matrix multiplication, and $\boldsymbol{E} = [\boldsymbol{E}_1, \dots, \boldsymbol{E}_m]$.

Consider a linear predictive model of $\boldsymbol{y}$ of the form $f(\boldsymbol{U}) = \boldsymbol{U\beta} + b$, where $\boldsymbol{\beta}$ are the coefficients, and $b$ is the intercept. To find latent components that encompass patterns in the input data and are predictive of the outcome, we propose the following optimization objective:

$$\min_{\boldsymbol{U},\boldsymbol{V},\boldsymbol{\beta}} \quad ||\boldsymbol{y} - \boldsymbol{U\beta}||_2^2 + \lambda||\boldsymbol{X} - \boldsymbol{UV}^T||_F^2 + \gamma||\boldsymbol{\beta}||_2^2$$

$$\text{s.t. } \boldsymbol{U}^T\boldsymbol{U} = \boldsymbol{I},\ \boldsymbol{V} = \boldsymbol{V} \circ \boldsymbol{S}, \tag{3}$$

where $||\cdot||_F$ denotes the Frobenius norm, $||\cdot||_2$ is the L2 norm, and $\lambda$ and $\gamma$ are hyperparameters. The first term is a prediction loss, the second term is a reconstruction loss, and the last term regularizes the coefficients of the predictive model. The component matrix $\boldsymbol{U}$ is constrained to be column-wise orthogonal to find non-redundant and complementary components. The loading matrix $\boldsymbol{V}$ is designed to obey the block-wise sparsity structure in $\boldsymbol{S}$. While the main goal is to find components that are predictive, the reconstruction loss serves as regularization to ensure meaningful components are found instead of noise, which is crucial in high-dimensional problems with high noise levels. We included L2 regularization rather than the sparsity inducing L1 because latent components are already compact representations of the input.

For classification, the hinge loss can be adopted to replace the prediction loss in the above regression setting, $\mathcal{L}_{hinge} = \sum_{i=1}^{n}\left(1 - \left(\boldsymbol{u}_i^T\boldsymbol{\beta} + b\right)y_i\right)_+^2$, where $(\cdot)_+ = \max(\cdot, 0)$. Since $y_i \in \{-1,1\}$, we have $1 - \left(\boldsymbol{u}_i^T\boldsymbol{\beta} + b\right)y_i = y_i(y_i - (\boldsymbol{u}_i^T\boldsymbol{\beta} + b))$. Let $z_i = y_i - \left(\boldsymbol{u}_i^T\boldsymbol{\beta} + b\right)$. We have the following optimization problem:

$$\min_{\boldsymbol{U},\boldsymbol{V},\boldsymbol{\beta}} \ (\boldsymbol{y} \circ \boldsymbol{z})_+^2 + \lambda||\boldsymbol{X} - \boldsymbol{UV}^T||_F^2 + \gamma||\boldsymbol{\beta}||_2^2$$

$$\text{s.t. } \boldsymbol{U}^T\boldsymbol{U} = \boldsymbol{I},\ \boldsymbol{V} = \boldsymbol{V} \circ \boldsymbol{S},\ \boldsymbol{z} = \boldsymbol{y} - (\boldsymbol{U\beta} + b) \tag{5}$$

## 3.2 *Optimization algorithm for the complete-modality case*

This section proposes an optimization algorithm to estimate parameters of MMFL for the classification case. The algorithm for the regression case can be found in Appendix A. The objective function in (5) is convex with respect to each of the parameters $\boldsymbol{U}, \boldsymbol{V}, \boldsymbol{\beta}$ while holding others constant. Thus, Alternating Minimization (AM) can be used to iteratively estimate the parameters. Since each of the coordinate-wise minimization problems is convex and have unique solution, AM is guaranteed to converge to a stationary point (Bertsekas, 2016).

We incorporate the third constraint via Augmented Lagrangian:

$$L(\boldsymbol{U}, \boldsymbol{V}, \boldsymbol{\beta}, b, \boldsymbol{q}) = (\boldsymbol{y} \circ \boldsymbol{z})_+^2 + \lambda \left|\left|\boldsymbol{X} - \boldsymbol{U}\boldsymbol{V}'\right|\right|_F^2 + \gamma \left|\left|\boldsymbol{\beta}\right|\right|_2^2 + \Phi(\boldsymbol{q}, \boldsymbol{z} - \boldsymbol{y} + (\boldsymbol{U}\boldsymbol{\beta} + b)), \tag{6}$$

where $\Phi(\boldsymbol{q}, \Delta) = \frac{\mu}{2}\left|\left|\Delta\right|\right|_F^2 + \langle \boldsymbol{q}, \Delta \rangle$ and $\boldsymbol{q}$ is the Lagrange multiplier. Taking the gradient with respect to each term in (6), we obtain a closed form updating equation for each term:

$$\begin{aligned}
\boldsymbol{V} &= \boldsymbol{X}^T \boldsymbol{U} \\
\boldsymbol{U} &= \left(\lambda \boldsymbol{X}\boldsymbol{V} + \tfrac{\mu}{2}(\boldsymbol{y} - b - \boldsymbol{z} - \tfrac{\boldsymbol{q}}{\mu})\boldsymbol{\beta}^T\right)\left(\lambda \boldsymbol{V}^T\boldsymbol{V} + \tfrac{\mu}{2}\boldsymbol{\beta}\,\boldsymbol{\beta}^T\right)^{-1} \\
\boldsymbol{z} &= \mathrm{I}_{\{\boldsymbol{y}\circ\boldsymbol{s}>\boldsymbol{0}\}} \circ \frac{\boldsymbol{s}}{\left(1+\frac{2}{\mu}\right)} + \mathrm{I}_{\{\boldsymbol{y}\circ\boldsymbol{s}\le\boldsymbol{0}\}} \circ \boldsymbol{s}\,, \text{ where } \boldsymbol{s} := \boldsymbol{y} - \boldsymbol{U}\boldsymbol{\beta} - b - \frac{\boldsymbol{q}}{\mu} \qquad (7) \\
\boldsymbol{\beta} &= \left(\frac{\mu}{2\gamma+\mu}\right)\left[\boldsymbol{U}^T\left(\boldsymbol{y} - b - \frac{\boldsymbol{q}}{\mu} - \boldsymbol{z}\right)\right] \\
b &= \frac{1}{n}\left(\boldsymbol{y} - \boldsymbol{z} - \boldsymbol{U}\boldsymbol{\beta} - \frac{\boldsymbol{q}}{\mu}\right)^T \boldsymbol{1}_n, \\
\boldsymbol{q} &\leftarrow \boldsymbol{q} + \mu(\boldsymbol{z} - \boldsymbol{y} + \boldsymbol{U}\boldsymbol{\beta} + b).
\end{aligned}$$

The constraints for orthogonality and block-wise sparsity are enforced at each iteration:

$$\begin{aligned}
&\boldsymbol{V} \leftarrow \boldsymbol{V} \circ \boldsymbol{S} \\
&\boldsymbol{U} \leftarrow \boldsymbol{L}\boldsymbol{R}^T \text{ from } \boldsymbol{U} = \boldsymbol{L}\boldsymbol{\Sigma}\boldsymbol{R}^T,
\end{aligned}$$

where the second line is the Singular Value Decomposition (SVD) solution to $\min_{\widehat{\boldsymbol{U}}} \left|\left|\boldsymbol{U} - \widehat{\boldsymbol{U}}\right|\right| \; s.t.\, \widehat{\boldsymbol{U}}^T\widehat{\boldsymbol{U}} = \boldsymbol{I}$. The proof can be found in Appendix B. The entire procedure for solving the MMFL optimization in (5) is summarized in Algorithm 1.

**Algorithm 1. Solving MMFL for classification problems with complete-modality data**

**Input:** Dataset $\{\{\boldsymbol{X}_1 \dots \boldsymbol{X}_m\}, \boldsymbol{y}\}$, structure matrix $\boldsymbol{S}$, stopping criterion $\epsilon$, hyperparameters $\gamma, \lambda$.
**Output:** Solutions for $\boldsymbol{U}, \boldsymbol{V}, \boldsymbol{\beta}$

1. **Initialize:** $\boldsymbol{U}^{(0)} \leftarrow \boldsymbol{L}\boldsymbol{R}^T$ from SVD of $\boldsymbol{X} = \boldsymbol{L}\boldsymbol{\Sigma}\boldsymbol{R}^T$
2. **Repeat**
   a. $k \leftarrow k + 1$;
   b. $\boldsymbol{V}^{(k)} = \boldsymbol{X}^T\boldsymbol{U}^{(k-1)}$
   c. $\boldsymbol{V}^{(k)} = \boldsymbol{V}^{(k)} \circ \boldsymbol{S}$
   d. $\boldsymbol{\beta}^{(k)} = \left(\frac{\mu}{2\gamma+\mu}\right)\left[\boldsymbol{U}^{(k-1)^T}\left(\boldsymbol{y} - b^{(k-1)} - \frac{\boldsymbol{q}^{(k-1)}}{\mu} - \boldsymbol{z}^{(k-1)}\right)\right]$
   e. $b^{(k)} = \frac{1}{n}\left(\boldsymbol{y} - \boldsymbol{z}^{(k-1)} - \boldsymbol{U}^{(k-1)}\boldsymbol{\beta}^{(k)} - \frac{\boldsymbol{q}^{(k-1)}}{\mu}\right)^T \mathbf{1}_n$
   f. $\boldsymbol{z}^{(k)} = \mathrm{I}\{\boldsymbol{y} \circ \boldsymbol{s} > \mathbf{0}\} \circ \frac{\boldsymbol{s}}{\left(1+\frac{2}{\mu}\right)} + \mathrm{I}\{\boldsymbol{y} \circ \boldsymbol{s} < \mathbf{0}\} \circ \boldsymbol{s}$ where $\boldsymbol{s} = \boldsymbol{y} - \boldsymbol{U}^{(k-1)}\boldsymbol{\beta}^{(k)} - b^{(k)} - \frac{\boldsymbol{q}^{(k-1)}}{\mu}$
   g. $\boldsymbol{U}^{(k)} = \left(\lambda\boldsymbol{X}\boldsymbol{V}^{(k)} + \frac{\mu}{2}\left(\boldsymbol{y} - b^{(k)} - \boldsymbol{z}^{(k)} - \frac{\boldsymbol{q}^{(k-1)}}{\mu}\right)\boldsymbol{\beta}^{(k)^T}\right)\left(\lambda(\boldsymbol{V}^{(k)})^T\boldsymbol{V}^{(k)} + \frac{\mu}{2}\boldsymbol{\beta}^{(k)}\,(\boldsymbol{\beta}^{(k)})^T\right)^{-1}$
   h. $\boldsymbol{U}^{(k)} = \boldsymbol{L}\boldsymbol{R}^T$ from $\boldsymbol{U}^{(k)} = \boldsymbol{L}\boldsymbol{\Sigma}\boldsymbol{R}'$
   i. $\boldsymbol{q}^{(k)} = \boldsymbol{q}^{(k-1)} + \mu\left(\boldsymbol{z}^{(k)} - \boldsymbol{y} + \boldsymbol{U}^{(k)}\boldsymbol{\beta}^{(k)} + b^{(k)}\right)$.
3. **until** $L\left(\boldsymbol{U}^{(k)}, \boldsymbol{V}^{(k)}, \boldsymbol{\beta}^{(k)}, b^{(k)}, \boldsymbol{q}^{(k)}\right) - L\left(\boldsymbol{U}^{(k-1)}, \boldsymbol{V}^{(k-1)}, \boldsymbol{\beta}^{(k-1)}, b^{(k-1)}, \boldsymbol{q}^{(k-1)}\right) \le \epsilon$

Given a new test sample, $\boldsymbol{x}^*$, we first obtain the latent components for the test sample by

$$\widehat{\boldsymbol{u}^*} = \underset{\boldsymbol{u}^*}{\operatorname{argmin}}\, \lambda \left|\left|\boldsymbol{x}^* - \boldsymbol{u}^*\widehat{\boldsymbol{V}}^T\right|\right|_F^2 = (\lambda\boldsymbol{x}^*\widehat{\boldsymbol{V}})\left(\lambda\widehat{\boldsymbol{V}}^T\widehat{\boldsymbol{V}} + \widehat{\boldsymbol{\beta}}\,\widehat{\boldsymbol{\beta}}^T\right)^{-1} \quad (11)$$

The prediction for the test sample is then $\hat{y} = \boldsymbol{u}^*\widehat{\boldsymbol{\beta}} + \hat{b}$.

## 3.3 *Optimization algorithm for the incomplete-modality case*

In the incomplete-modality case, the objective function in (5) can be modified to

$$\min_{\boldsymbol{U},\boldsymbol{V},\boldsymbol{\beta}} (\boldsymbol{y} \circ \boldsymbol{z})_+^2 + \lambda \sum_{k=1}^{m} \left|\left|\mathrm{P}_{\boldsymbol{O}_k}\left(\boldsymbol{X}_k - \boldsymbol{U}\,\boldsymbol{V}_k^T\right)\right|\right|_F^2 + \gamma\left|\left|\boldsymbol{\beta}\right|\right|_2^2, \quad (8)$$

where $\mathrm{P}_{\boldsymbol{O}_k}(\boldsymbol{X}_k) = \boldsymbol{X}_k \circ \boldsymbol{O}_k$ and $\boldsymbol{O}_k$ is a binary diagonal matrix with its $i$-th element indicating the existence of the $k$-th modality data for the $i$-th sample, i.e., value of 1 when the modality $k$ is available for subject $i$, and 0 otherwise. Under this formulation, the reconstruction loss is minimized over entries of non-missing modalities, while the prediction loss is optimized over all samples.

The closed form solution of $\boldsymbol{U}$ becomes

$$\boldsymbol{u}_i = \left(\lambda \sum_{k\in\Omega_i} \boldsymbol{x}_{i,k}\boldsymbol{V}_k + (\boldsymbol{y} - b - \boldsymbol{z} - \tfrac{\boldsymbol{q}}{\mu})\boldsymbol{\beta}^T\right)\left(\lambda \sum_{k\in\Omega_i} \boldsymbol{V}_k^T\boldsymbol{V}_k + \boldsymbol{\beta}\,\boldsymbol{\beta}^T\right)^{-1}, \quad (9)$$

where $\Omega_i \subseteq \{1, \dots, m\}$ is the set of non-missing modalities for sample $i$. Note that during each update, the entire latent component matrix is estimated for both complete and incomplete modality samples, and component orthogonality is subsequently enforced across all samples.

When a modality is missing, its individual component is not observed, but its globally joint and partially joint components exist in other observed modalities and could be recovered from the decomposition. Thus, we can obtain a pseudo-reconstruction of missing modalities by

$$\widehat{\boldsymbol{X}}_{mis} = \widehat{\boldsymbol{U}}_{mis}\widehat{\boldsymbol{V}}^T. \tag{10}$$

This is not intended to be an imputation. Rather, it is a partial reconstruction of $\boldsymbol{X}_{mis}$ based on the information that the model was able to learn from other observed modalities via internal transfer learning.

After this step, Algorithm 1 can be used to re-estimate the parameters. Parameter estimation and pseudo-reconstruction can be iterated until convergence. This procedure is described in Algorithm 2. In our experiments, we observed that initializing $\boldsymbol{V}, \boldsymbol{\beta}$ using complete-modality samples usually leads to fast convergence within a few iterations.

**Algorithm 2. Solving MMFL for classification problems with incomplete-modality data**

**Input:** Dataset $\{\{\boldsymbol{X}_1 \dots \boldsymbol{X}_m\}, \boldsymbol{y}\}$, structure matrix $\boldsymbol{S}$, stopping criterion $\epsilon$, hyperparameters $\gamma, \lambda$.
**Output:** Solutions for $\boldsymbol{U}, \boldsymbol{V}, \boldsymbol{\beta}$

1. **Initialize:** $\boldsymbol{V}^{(0)}$, $\boldsymbol{\beta}^{(0)}$
2. **Repeat**
   a. $k \leftarrow k + 1$;
   b. Parameter estimation: $\boldsymbol{U}^{(k)}, \boldsymbol{V}^{(k)}, \boldsymbol{\beta}^{(k)} \leftarrow Algorithm1(\{\{\boldsymbol{X}_1 \dots \boldsymbol{X}_m\}, \boldsymbol{y}\})$ with (9)
   c. Pseudo-reconstruction: $\widehat{\boldsymbol{X}}_{mis}^{(k)} = \boldsymbol{U}_{mis}^{(k)}(\boldsymbol{V}^{(k)})^T$

4. **until** $L(\boldsymbol{U}^{(k)}, \boldsymbol{V}^{(k)}, \boldsymbol{\beta}^{(k)}, b^{(k)}, \boldsymbol{q}^{(k)}) - L(\boldsymbol{U}^{(k-1)}, \boldsymbol{V}^{(k-1)}, \boldsymbol{\beta}^{(k-1)}, b^{(k-1)}, \boldsymbol{q}^{(k-1)}) \le \epsilon$

Given a new test sample, $\boldsymbol{x}^*$, the latent components for the test sample $\boldsymbol{u}^*$ is obtained by

$$\boldsymbol{u}^* = \underset{\boldsymbol{u}^*}{\operatorname{argmin}} \lambda \sum_{k=1}^{m} \left\| \mathrm{P}_{\boldsymbol{O}_k}\left( \boldsymbol{x}_k^* - \boldsymbol{u}^* \widehat{\boldsymbol{V}}_k^T \right) \right\|_F^2 = \left(\lambda \sum_{k \in \Omega_i} \boldsymbol{x}_k^* \widehat{\boldsymbol{V}}_k\right) \left(\lambda \sum_{k \in \Omega_i} \widehat{\boldsymbol{V}}_k^T \widehat{\boldsymbol{V}}_k + \widehat{\boldsymbol{\beta}}\widehat{\boldsymbol{\beta}}^T\right)^{-1} \tag{12}$$

The prediction for the test sample is then $\hat{y} = \boldsymbol{u}^*\widehat{\boldsymbol{\beta}} + \hat{b}$. This formulation allows test samples to have incomplete modalities. Nevertheless, it is expected that the performance for incomplete-modality samples would drop because their latent components are estimated with less supervision.

## 3.4 *Rank selection*

Rank selection is an inherently challenging problem for multi-modal latent variable models. Exhaustive search all possible rank combinations across the different components can be computationally infeasible. Different approaches to narrow down the search space have been proposed in the literature. For example, Lock et al. (Lock et al., 2013) first selected ranks of the shared component, then determined ranks of each individual component conditional on the shared. Gaynanova and Li (Gaynanova & Li, 2019) used a bi-cross-validation-based approach to simultaneously select the best structure from a fixed set. Inspired by previous works, we propose two approaches for rank selection of MMFL:

(a) Sequential approach: The user pre-sets an order for which component ranks will be determined. Initially, all ranks are set to 0. Starting for the first component, increase its rank from 0, 1, 2, …, until the performance metric does not improve or the maximum possible rank $r_{max} = \min(n, p)$ is reached. Fix the rank for this component. Move onto the next component and repeat the process until all component ranks are determined.

(b) Incremental approach: All ranks are initialized to 0. At each iteration, select one component for which increasing its rank by one while keeping the ranks of other components fixed leads to the best performance metric improvement. Increase the rank for the selected component. Repeat the process until no component is selected or $r_{max}$ is reached for all components.

In applications where some components should be prioritized over others, for example, finding globally joint components across all modalities is of great interest to the user, the sequential approach should be used. Otherwise, the incremental approach should be used as it provides less biased rank selection.

A top-down approach can be used to approximate a reasonable range of values for component ranks. This can be done by running MMFL using the maximum number of ranks for every component, then visualizing the magnitude of loadings averaged for each component and counting the number of components with non-trivial loadings. This estimate can serve as a starting point and speed up the rank selection process. Similarly, once rank selection is finalized, the loading matrices can be visualized to ensure all included ranks are non-trivial and ranks can be re-adjusted as necessary.

Remark: It is possible that no unique solution for rank selection optimizes the MMFL formulation. Mathematically, a joint component can be represented as multiple individual components, and multiple individual components may "sneak" into one joint component yielding the same reconstruction and prediction loss. Thus, different rank selection solutions may exhibit similar predictive performance. Selecting appropriate ranks can still be of interest for interpretability. The MMFL model formulation ensures the selected ranks are complementary and non-overlapping. We acknowledge that rank selection is a challenging problem, and the proposed approaches are not necessarily optimal. Future research can be conducted to explore more along this direction.

## 3.5 *Complementary and consensus principles*

In this section, we provide some discussion of the design rationale behind MMFL, which underpins its success. We discuss two fundamental principles that have been followed by most existing methods in the multi-modal learning literature: *consensus* and *complementary* principles (Xu et al., 2013).

- *Consensus principle* aims to maximize the agreement of models trained on multiple modalities. Co-training is a multi-modal learning strategy that fully relies on the consensus principle. In fact, theoretical analysis on error bounds for co-training showed that the probability of disagreement of two independent hypotheses developed using different features upper bounds the error rate of either hypotheses (Dasgupta et al., 2001). Besides prediction-level agreement, maximizing feature representation agreement across modalities, i.e. extracting the globally joint components, is another form of consensus. Either approach assumes that all modalities are somewhat connected, either in the feature space or in their connection with the outcome.
- *Complementary principle* aims to maximize diversity of inputs to fully explore the hypothesis space. It is assumed that each modality contains some knowledge that other modalities do not have and these modality-specifics are non-conflicting. Models that extracting non-overlapping modality-specific information follow the complementary principle. Orthogonality constraints and Hilbert Schmidt Independent Criterion (HSIC) (J. Li et al., 2020)-based penalties are typically added to the model's objective function to induce diversity.

Only following complementary principle will ignore the connection of multiple modalities, such as the case of SUM-PCA (Smilde et al., 2003), while only following the consensus principle does not make full use of the complementary information in multiple datasets, such as the case of CMLS (T. Zhou et al., 2020). Thus, both principles are important for effective multi-modal learning. The design of MMFL simultaneously considers the two principles, decomposing multi-modal data into different levels of joint and individual components that are orthogonal to each other.

## 4. Simulation Study

In this section, we assess the performance of MMFL using simulation data, in comparison with several competing methods, under complete-modality and incomplete-modality settings.

### 4.1 *Simulation setup*

We first sampled the response variable as $\boldsymbol{y} \sim Bernoulli(0.5)$. We generated $\boldsymbol{U}$ conditional on $\boldsymbol{y}$ as

$$\boldsymbol{u}_i \sim \begin{cases} U(0,1) & if\ y_i = 0 \\ U(\Delta, 1+\Delta) & if\ y_i = 1 \end{cases}$$

where $\Delta$ quantifies the class separability of the data. $\boldsymbol{U}$ is orthogonalized to represent non-overlapping sources of information. Then, we generated the loadings $\boldsymbol{V} \sim U(-1,1)$ and set $\boldsymbol{V} = \boldsymbol{V} \circ \boldsymbol{S}$, where $\boldsymbol{S}$ encodes a structure that contains globally joint, partially joint, and individual components, each with rank $r_1 = \cdots r_7 = 3$. The feature matrix is computed $\boldsymbol{Z} = \boldsymbol{U}\boldsymbol{V}'$ and Gaussian noise is added as $\boldsymbol{X}_i = \boldsymbol{Z}_i + \boldsymbol{E}_i, i = 1 \ldots m$, $\boldsymbol{E}_i \sim N(0, {\sigma_i}^2)$. We set $\sigma_i$ such that signal-to-noise ratio (SNR) $\frac{||\boldsymbol{Z}_i||_F^2}{\mathrm{E}\left(||\boldsymbol{E}_i||_F^2\right)} = \frac{||\boldsymbol{Z}_i||_F^2}{({\sigma_i}^2 \mathrm{np_i})} = i$ for modality $i$, i.e. the first modality have the highest SNR of 1, the second modality have SNR of 2, and so on. Note that we purposely constructed datasets with relatively high noise to create challenging scenarios that can show performance difference in the models.

We simulated datasets with three modalities each with 100 features. We generated 200 training samples and 200 test samples. A unique trait of MMFL is the capability to handle incomplete multi-modal data in both training and test phases. Thus, we simulated incomplete modalities in both training and test datasets. In the training dataset, 0%, 20% and 40% of the rows were randomly and independently masked out from each of the three modalities. To simulate missing data in the test phase, multiple versions of the test dataset were created by removing different subsets of modalities from all samples, generating different sample cohorts with distinct data availabilities. $\Delta$ is set to $0.50$ for incomplete modality setting experiments and $0.25$ for complete modality experiments to simulate a more challenging scenario to compare the various methods.

Methods were compared in terms of average test accuracy, Area Under the Curve (AUC), and training time over 20 random replications. To compute accuracy, probabilistic predictions were converted to binary predictions based on the cut-off determined by Youden's index (Ruopp et al., 2008). For fair comparison of methods without being affected by the rank selection algorithm, ground truth ranks were used for all methods.

## 4.2 *Competing methods*

As the key innovation of MMFL is incorporating supervision from labels and decomposing multi-modal data into globally joint, partially joint, and individual components. We compared MMFL with a range of methods that possess related capabilities:

(a) JIVE (Lock et al., 2013), a classic unsupervised multi-modal model that decomposes data into globally joint and individual components;

(b) SLIDE (Gaynanova & Li, 2019), an unsupervised method that extracts globally joint, partially joint, and individual components from multi-view data;

(c) sJIVE (Palzer et al., 2022), a recent supervised extension of JIVE to extract predictive latent globally joint and individual components;

(d) CMLS (T. Zhou et al., 2020), a supervised method that extracts globally joint components and an ensemble of classifiers for prediction.

(e) IMLS (T. Zhou et al., 2020), the CMLS model extended to incomplete-modality settings.

For fair comparison, all selected methods are linear. Table 1 summarizes their capabilities.

Table 1. Comparison of MMFL and competing methods for their capabilities

| Method | Globally joint components | Partially joint components | Supervision from y | Incomplete modality | Rank selection |
|---|---|---|---|---|---|
| JIVE | X | | | | X |
| SLIDE | X | X | | | X |
| CMLS | X | | X | | |
| IMLS | X | | X | X | |
| sJIVE | X | | X | | |
| MMFL (proposed) | X | X | X | X | X |

We tuned MMFL's hyperparameters within the ranges $\lambda \in \{0.1,1,10\}$, $\gamma \in \{0.001,0.01\}$; the hyperparameter of sJIVE within the ranges $\eta \in \{0.1 \dots 0.9\}$; and those of CMLS/IMLS within the ranges $\{\gamma, \beta, \lambda\} \in \{10^{-3}, 10^{-2}, \dots, 10^{3}\}$, $h \in \{10,20\}$ as suggested by the authors. All hyperparameters were tuned via grid search and five-fold cross validation (CV) on one training set. For the two unsupervised methods, a logistic regression model was trained on the concatenated components to produce classification outputs.

## 4.3 *Complete modality case*

This experiment compares different methods in the complete modality setting to demonstrate the advantage of MMFL for having supervision from labels and for decomposing data into globally shared, locally shared, and individual components. The results for a challenging high-noise classification setting ($\Delta = 0.25$) is summarized in Tables 2. As expected, unsupervised methods performed poorly because of the high noise levels in the dataset and lack of supervision to find predictive latent components. SLIDE had significantly better performance than JIVE because of its capability to identify partially joint components. CMLS's performance was poor, likely because it only focused on extracting information common in all modalities. sJIVE also had poor performance because it ignored predictive information contained in partially joint components. MMFL significantly outperformed other models in accuracy and training time, showing the advantage of having supervision and encoding the full modality structure.

Table 2. Model performance in the complete modality setting.

| | Test AUC | Test accuracy | Training Time (s) |
|---|---|---|---|
| SLIDE | 0.685 (0.037) | 0.659 (0.034) | 0.590 (0.181) |
| JIVE | 0.504 (0.046) | 0.527 (0.041) | 13.664 (6.434) |
| sJIVE | 0.506 (0.051) | 0.527 (0.042) | 2.664 (1.152) |
| CMLS | 0.560 (0.045) | 0.566 (0.041) | 11.966 (3.085) |
| MMFL (proposed) | **0.818 (0.037)** | **0.758 (0.037)** | **0.534 (0.152)** |

## 4.4 *Incomplete modality case*

This experiment assesses MMFL's performance in the incomplete multi-modal data setting. As an additional competing method, we trained a MMFL model using only complete modality samples (training with less samples). Another intuitive competing method for test samples with incomplete modality is to train one model specific to each cohort and predict using modalities that are available at test time (training with less modalities). Among the competing methods mentioned in Section 4.2, only IMLS is designed to handle incomplete modality datasets and was included in this experiment.

As a performance upper bound, MMFL classified the complete modality test cohort with 0.944 AUC and 0.876 accuracy when all training samples are complete modality. Having missing rate of 0%, 20%, 40% in training, the performance of MMFL slightly decreased to 0.919 AUC and 0.848 accuracy, which still outperforms IMLS and the MMFL model trained using only complete modality samples (Table 3). When predicting for test cohorts with incomplete modalities, MMFL's test AUC significantly dropped to 0.787, 0.776, and 0.768, the later had relatively lower performance likely because the modality with the lower noise ($X_3$) was dropped instead of the modality with higher noise ($X_1$). Despite the performance downgrade, MMFL still outperformed IMLS and cohort-specific models trained using only the available modalities.

Table 3. Model performance in the incomplete modality setting.

| Test cohort | Model | AUC | Accuracy |
|---|---|---|---|
| Complete modality at test time | | | |
| Cohort with $(X_1, X_2, X_3)$ | IMLS | 0.659 (0.045) | 0.632 (0.034) |
| | MMFL trained on complete samples | 0.873 (0.033) | 0.808 (0.033) |
| | MMFL trained on all samples | **0.919 (0.029)** | **0.848 (0.039)** |
| Incomplete modality at test time | | | |
| Cohort with $(X_1, X_2)$ | IMLS | 0.635 (0.044) | 0.617 (0.036) |
| | MMFL trained on $(X_1, X_2)$ | 0.743 (0.045) | 0.695 (0.036) |
| | MMFL trained on all samples | **0.768 (0.047)** | **0.713 (0.039)** |
| Cohort with $(X_1, X_3)$ | IMLS | 0.637 (0.042) | 0.616 (0.037) |
| | MMFL trained on $(X_1, X_3)$ | 0.735 (0.039) | 0.687 (0.032) |
| | MMFL trained on all samples | **0.776 (0.035)** | **0.718 (0.032)** |
| Cohort with $(X_2, X_3)$ | IMLS | 0.638 (0.048) | 0.617 (0.036) |
| | MMFL trained on $(X_2, X_3)$ | 0.738 (0.046) | 0.686 (0.034) |
| | MMFL trained on all samples | **0.787 (0.045)** | **0.725 (0.039)** |

## 5. Case study: Early Prediction of Alzheimer's Disease

AD is a devastating neurodegenerative disorder that gradually destroys memory and cognitive abilities, currently affecting 6.7 million people aged 65 and older in the U.S ("2023 Alzheimer's Disease Facts and Figures," 2023). Existing disease-modifying treatment strategies to slow down the disease progression rely on early intervention (Yiannopoulou & Papageorgiou, 2020). Thus, early prediction of which patients will progress to AD, in particular, during the MCI stage when patients show noticeable signs of memory loss and cognitive decline, is crucial for planning time-effective treatments. In this section, we present an application of using multimodality neuroimaging data and genetic data for the early prediction of AD for MCI patients.

## 5.1 *Data preparation*

We used data from the Alzheimer's Disease Neuroimaging Initiative (ADNI). ADNI (http://adni.loni.ucla.edu) is a cooperative study that contains data from over 50 universities and medical centers in the United States and Canada to study the progression of AD (Mueller et al., 2005). The primary goal of ADNI has been to test whether magnetic resonance imaging (MRI), position emission tomography (PET), genetic markers, clinical and neuropsychological assessments can be combined to measure the progression of MCI and early AD.

We downloaded 1318 samples from MCI patients in the ADNI database. Three data modalities were included: T1-weighted volumetric MRI, fluorodeoxyglucose (FDG) PET, and genome-wide association study (GWAS) single nucleotide polymorphism (SNP) arrays. All samples have MRI available (479 converters and 839 non-converters according to a three-year conversion time window). A subset of 1002 samples has PET (388 converters and 614 non-converters) and a subset of 532 samples has SNP (270 converters and 262 non-converters). Note that the 1318 samples were from 563 patients since each patient may have multiple visits for data collection. Here we treated each visit as a sample to increase sample size. To avoid potential risk of overfitting by using this strategy, we performed train/test split by patients such that no patient has samples in both training and test sets and the proposition of converters is maintained similar in both sets. Since majority of competing methods cannot handle incomplete modalities at test time, we constructed the test set by sampling 20% of the complete modality samples. All models were trained and tested on 20 random replications. Age, sex, and apolipoprotein E (APOE) gene status were included as additional predictors in all models.

The MRI data was processed by FreeSurfer v7.1 to obtain volumetric measures following standard procedures (Desikan et al., 2006; Fischl, 2004). PET was processed by a PET Unified Pipeline to obtain regional standardized uptake value ratios (SUVR) measurements for FreeSurfer defined regions. We included volumetric measures for 68 cortical and 14 sub-cortical regions of interest (ROI), totalling 82 MRI and 82 PET features or imaging phenotypes for each sample.

Quality control was performed on SNP data in PLINK v1.90 using the following criteria based on standard protocols (Marees et al., 2018): genotype call rate ≥ 95%, minor allele frequency (MAF) ≥ 10%, Hardy Weinberg Equilibrium (HWE) test P ≤ 1.0E-6, heterozygosity rates within ±3 SD of the sample, cryptic relatedness rate ≤ 0.2, and a platform-specific recommended quality control score of 0.15. After quality control, 454,442 out of 620,901 SNPs remained. The association of genotypes with AD conversion was tested using PLINK'S genome association analysis toolset (Purcell et al., 2007) with a significance level set to $P < 5 \times 10^{-4}$ to ensure a relatively large number of SNPs are included. We additionally included SNPs that were reported to have association with AD according to a recent large-scale study (Bellenguez et al., 2022). This combination of knowledge- and data-driven strategy selected 208 SNPs which were used as genotype features for downstream models.

## 5.2 *Case study prediction results*

We compared MMFL with the competing methods listed in Section 4.2. Majority of these competing methods do not have built-in designs to handle incomplete modality data. In order to still compare with them and validate the benefit of leveraging incomplete modality samples, we trained SLIDE, JIVE, sJIVE using only complete modality samples (about one third of training samples). The same rank selection method was applied to all models for fair comparison. In this case study, sequential rank selection that determines ranks first for globally joint component followed by partially joint and individual components showed better results for all methods compared to incremental rank selection.

The classification results are summarized in Table 4. All supervised models significantly outperformed the unsupervised models, demonstrating the benefit for using supervision from the label to find predictive latent components compared to a two-step approach. Among the unsupervised methods, SLIDE, which enforced more modality structure for the decomposition, significantly outperformed JIVE. MMFL outperformed sJIVE and IMLS, showing the advantage of identifying partially joint components. The close performance of IMLS with sJIVE and MMFL suggest that a few globally joint components extracted across all modalities was sufficient to capture majority of predictive information in this dataset. It is also worth noting that the training time advantage of MMFL observed in the simulation studies no longer applies because MMFL has a greater number of ranks to be determined, therefore more iterations of the rank selection algorithm need to be run.

Table 4. Test performance on ADNI data.

| | Test AUC | Test accuracy | Training time (s) |
|---|---|---|---|
| SLIDE | 0.751 (0.039) | 0.700 (0.036) | 2.787 (0.522) |
| JIVE | 0.496 (0.082) | 0.562 (0.464) | 32.646 (6.029) |
| sJIVE | 0.812 (0.058) | 0.746 (0.053) | 14.886 (3.948) |
| IMLS | 0.802 (0.060) | 0.728 (0.070) | 65.298 (3.927) |
| MMFL (proposed) | **0.826 (0.065)** | **0.766 (0.060)** | 46.193 (4.191) |

## 5.3 *Interpretation of latent components*

We visualized the loading matrices to better understand the composition of the components found by each of the three best models in Figure 1. In each plot, vertical axis are the concatenated globally joint, locally joint, and individual component loading matrices, as applicable, and the horizontal axis are the features. For simpler visual, only the top 5$^{th}$ and bottom 95$^{th}$ values within each component block (globally joint, partially joint, individual) are shown, which we consider the most predictive features. Loadings were multiplied by the sign of the respective coefficient to convert them to AD risk-prone loadings. For example, negative loadings (colored in blue) imply that lower feature values are associated with higher risk for AD.

All three models extracted a few globally joint components. However, the composition of their globally joint components is very different. The components extracted by IMLS were highly correlated with each other as no orthogonality constraint was imposed in the latent space. Moreover, the globally joint components extracted by IMLS and some by sJIVE were not truly globally joint but partially joint component with contributing features from MRI and PET. In contrast, all modalities contributed to the globally joint components identified by MMFL, suggesting possible correlations between caudate, superior frontal, left precentral, and right parahippocampal regions with a series of genetic phenotypes. Interestingly, MMFL only identified partially joint components between MRI-PET and not between MRI-SNP or PET-SNP, implying that there is more correlation among the neuroimaging modalities and fewer correlation between neuroimaging and genetic data. Besides the joint components, five individual components were identified for PET, three for MRI, and one for SNP, suggesting that PET contained multiple sources of predictive information for predicting AD.

The top contributing imaging features in sJIVE and MMFL were blue, indicating that samples with reduced in volume would be predicted with higher probability of AD conversion. While volume reduction in hippocampus (Convit et al., 1997) is long known to be pronounced for individuals with high risk for dementia, other regions highlighted by MMFL such as amygdala, parahippocampus, middle and inferior temporal, frontal gyrus, lingual gyrus, and parietal regions have also been identified to play an important role in the development of AD or the disease's early symptoms in previous studies (L. Li et al., 2022; H. Wang et al., 2012).

Both MMFL and sJIVE found multiple genotypes to be predictive, whereas IMLS only identified one predictive SNP. The following top contributing SNPs in both MMFL's globally joint and individual components were known to be associated with AD risk factors: rs7255066, a SNP located in a renown neurological protein biomarker gene (PVR) (Hillary et al., 2019) and was previously found to be associated with AD (Guo et al., 2012); rs1417182, a SNP with significant association with obsessive compulsive disorder (Sampaio et al., 2010); and rs9390790, a SNP associated with epilepsy (Guo et al., 2012), the latter two being potential risk factors for AD (Wu et al., 2012; Zhang et al., 2022). The joint component additionally highlighted rs10851884, a SNP associated with major depressive disorder, another risk factor for AD (Ferentinos et al., 2014; Huang et al., 2020). Thus, we conclude that MMFL provided slightly more accurate predictions along and extracted more interpretable components than IMLS and sJIVE.

Figure 1. Visualization of loading matrices for the three best methods.

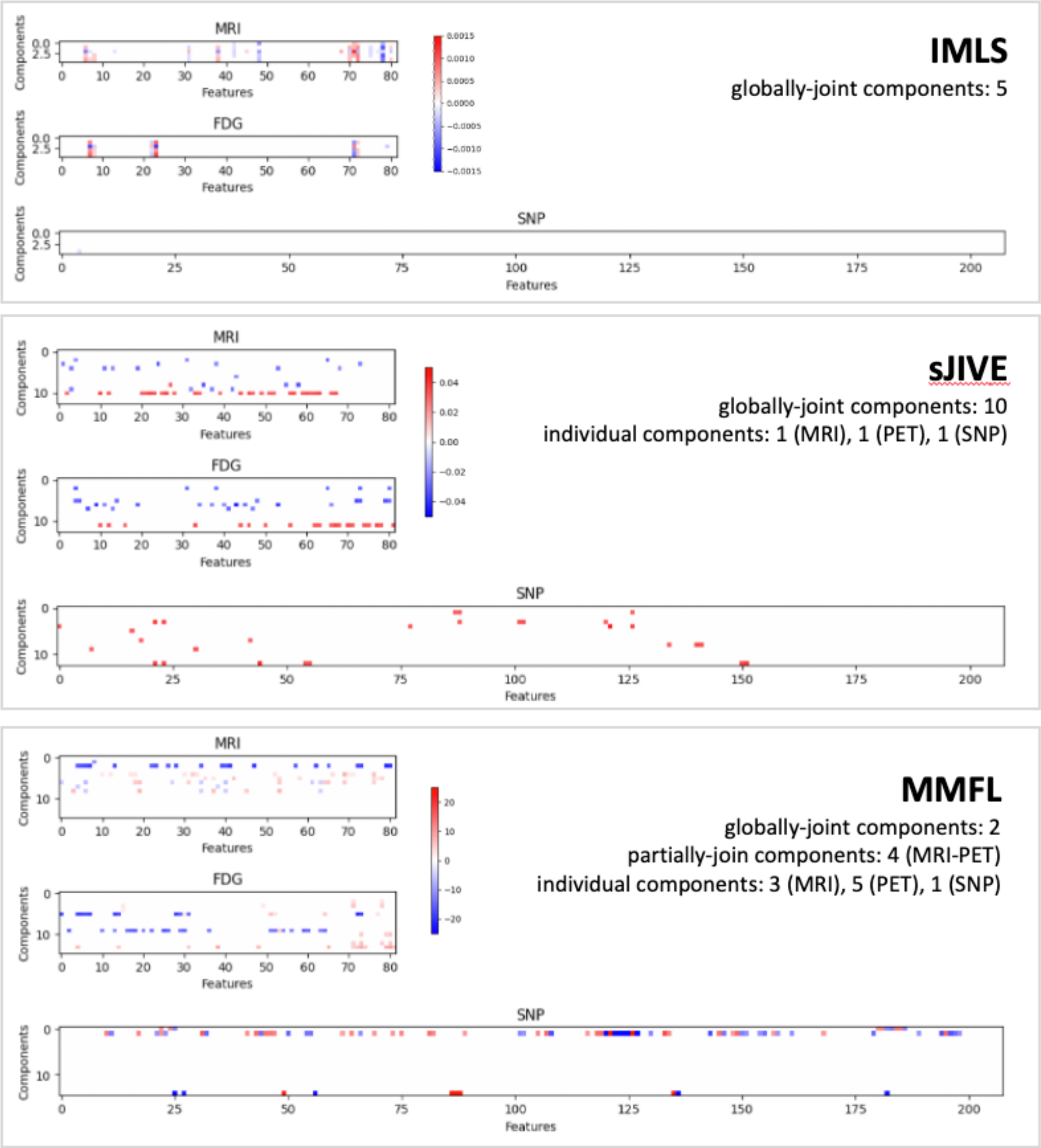

## 5. Conclusion

We proposed a novel supervised multi-modal fission learning (MMFL) model that simultaneously considers within-modality and cross-modality correlations to extract globally joint, partially joint, and individual components that are predictive of the outcome. Compared with existing methods, MMFL maximally reduces redundancy in multi-modal datasets, generating succinct, effective predictors. Another strength of MMFL is the ability to handle incomplete modality data in both training and test phases. We evaluated MMFL using simulation studies and a case study for early detection of AD.

MMFL showed better predictive performance compared to various multi-modal algorithms under both complete modality and incomplete modality settings. A limitation of the proposed method is the rank selection algorithm. While there is no unique solution to the rank selection problem, selecting appropriate ranks can affect model accuracy and interpretability. The proposed rank selection heuristics can be sensitive to noise and significantly slowed down the training time of MMFL. Furthermore, both the decomposition and prediction are based on linear models. A nonlinear extension of our model will improve its capability to model datasets with complex patterns.

## Acknowledgements

This research was supported by NIH grant 2R42AG053149-02A1 and NSF grant DMS-2053170. This research was also supported by NIH grants R01AG069453 and 30AG072980, the State of Arizona, and Banner Alzheimer's Foundation.

## Disclosure statement

The authors report there are no competing interests to declare.

## Data availability

The data used in this study can be downloaded from ADNI upon approval.

## References

Automatic citation updates are disabled. To see the bibliography, click Refresh in the Zotero tab.

### Appendix A: Regression formulation

The MMFL optimization problem for the complete modality case is

$$\min_{\boldsymbol{U},\boldsymbol{V},\boldsymbol{\beta}} \ ||\boldsymbol{y} - \boldsymbol{U}\boldsymbol{\beta}||_2^2 + \lambda||\boldsymbol{X} - \boldsymbol{U}\boldsymbol{V}^T||_F^2 + \gamma||\boldsymbol{\beta}||_2^2$$

$$\text{s.t. } \boldsymbol{U}^T\boldsymbol{U} = \boldsymbol{I}, \boldsymbol{V} = \boldsymbol{V} \circ \boldsymbol{S}\ . \tag{12}$$

Taking the gradient of (12) with respect to each term, we obtain the following update equations:

$$\boldsymbol{V} = \boldsymbol{X}^T\boldsymbol{U};$$

$$\boldsymbol{\beta} = (1+\gamma)^{-1}\boldsymbol{U}^T\boldsymbol{y};$$

$$\boldsymbol{U} = (\lambda\boldsymbol{X}\boldsymbol{V} + \boldsymbol{y}\boldsymbol{\beta}^T)(\lambda\boldsymbol{V}^T\boldsymbol{V} + \boldsymbol{\beta}\boldsymbol{\beta}^T)^{-1}; \tag{13}$$

The MMFL optimization problem for the incomplete modality case is

$$\min_{\boldsymbol{U},\boldsymbol{V},\boldsymbol{\beta}} ||\boldsymbol{y} - \boldsymbol{U}\boldsymbol{\beta}||_2^2 + \lambda \sum_{j=1}^{m} \left|\left|\mathrm{P}_{\boldsymbol{O}_j}\left(\boldsymbol{X}_j - \boldsymbol{U}\,\boldsymbol{V}^T{}_j\right)\right|\right|_F^2 + \gamma||\boldsymbol{\beta}||_2^2$$

$$\text{s.t. } \boldsymbol{U}^T\boldsymbol{U} = \boldsymbol{I}, \boldsymbol{V} = \boldsymbol{V} \circ \boldsymbol{S}\,. \quad (14)$$

The update equations for $\boldsymbol{U}$ becomes

$$\boldsymbol{u}_i = \left(\lambda \sum_{k\in\Omega_i} \boldsymbol{x}_{i,k}\boldsymbol{V}_k + y_i\boldsymbol{\beta}^T\right)\left(\lambda \sum_{k\in\Omega_i} \boldsymbol{V}_k^T\boldsymbol{V}_k + \boldsymbol{\beta}\,\boldsymbol{\beta}^T\right)^{-1}. \quad (15)$$

The rest of the algorithm follows the classification case.

## Appendix B: Orthogonalizing matrices using SVD

Given $\boldsymbol{U}$, we want to find a new $\widehat{\boldsymbol{U}}$ that is orthogonal and closest to $\boldsymbol{U}$, i.e.

$$\min_{\widehat{\boldsymbol{U}}} \left|\left|\boldsymbol{U} - \widehat{\boldsymbol{U}}\right|\right| \; s.t. \widehat{\boldsymbol{U}}^T\widehat{\boldsymbol{U}} = \boldsymbol{I}.$$

We have

$$\min_{\widehat{\boldsymbol{U}}} \left|\left|\boldsymbol{U} - \widehat{\boldsymbol{U}}\right|\right| = \min_{\widehat{\boldsymbol{U}}} < \boldsymbol{U} - \widehat{\boldsymbol{U}}, \boldsymbol{U} - \widehat{\boldsymbol{U}} > = ||\boldsymbol{U}||_F^2 + I - 2 < \widehat{\boldsymbol{U}}, \boldsymbol{U} >_F$$

$$= \max_{\widehat{\boldsymbol{U}}} < \widehat{\boldsymbol{U}}, \boldsymbol{U} >_F = \max_{\widehat{\boldsymbol{U}}} < \widehat{\boldsymbol{U}}, \boldsymbol{L}\boldsymbol{\Sigma}\boldsymbol{R}^T >_F = \max_{\widehat{\boldsymbol{U}}} < \boldsymbol{L}^T\widehat{\boldsymbol{U}}\boldsymbol{R}, \boldsymbol{\Sigma} >_F.$$

Since $\boldsymbol{L}^T\widehat{\boldsymbol{U}}\boldsymbol{R}$ is orthogonal and $\boldsymbol{\Sigma}$ is diagonal, maximum is achieved when $\boldsymbol{L}^T\widehat{\boldsymbol{U}}\boldsymbol{R} = \boldsymbol{I}$. Thus $\widehat{\boldsymbol{U}} = \boldsymbol{L}\boldsymbol{R}^T$.